\documentclass[letterpaper, 10 pt, conference]{ieeeconf}  % Comment this line out if you need a4paper

\IEEEoverridecommandlockouts                              % This command is only needed if 
\usepackage{multirow}
\usepackage{multicol}
\usepackage{graphicx}
\usepackage{subcaption}
\usepackage{makecell}
\usepackage{amsmath}
\usepackage{algorithm}
\usepackage{algpseudocode}
\usepackage{xcolor}
\usepackage{fancyhdr}

\fancypagestyle{firstpage}{%
  \lhead{\fontsize{6pt}{11pt}\selectfont This work has been submitted to the IEEE for possible publication. Copyright may be transferred without notice, after which this version may no longer be accessible.}
}

\title{\LARGE \bf
Learning with Chemical versus Electrical Synapses\\
Does it Make a Difference?
}

\author{M\'{o}nika Farsang$^{1*}$, Mathias Lechner$^{2}$, David Lung$^{1}$, Ramin Hasani$^{2}$, Daniela Rus$^{2}$, Radu Grosu$^{1}$% <-this % stops a space
\thanks{$^{1}$CPS, Technische Universität Wien (TU Wien), Austria}%
\thanks{$^{2}$CSAIL, Massachusetts Institute of Technology (MIT), MA, USA}%
\thanks{$^{*}${\tt\small monika.farsang@tuwien.ac.at}}%
\thanks{**This project has received funding from the European Union’s Horizon 2020 research and innovation programme under the Marie Skłodowska-Curie grant agreement No 101034277.}
}

\begin{document}

\maketitle
\thispagestyle{firstpage}

%\thispagestyle{empty}
%\pagestyle{empty}

%%%%%%%%%%%%%%%%%%%%%%%%%%%%%%%%%%%%%%%%%%%%%%%%%%%%%%%%%%%%%%%%%%%%%%%%%%%%%%%%
\begin{abstract}
Bio-inspired neural networks have the potential to advance our understanding of neural computation and improve the state-of-the-art of AI systems. Bio-electrical synapses directly transmit neural signals, by enabling fast current flow between neurons. In contrast, bio-chemical synapses transmit neural signals indirectly,  through neurotransmitters. Prior work showed that interpretable dynamics for complex robotic control, can be achieved by using chemical synapses, within a sparse, bio-inspired architecture, called Neural Circuit Policies (NCPs). However, a comparison of these two synaptic models, within the same architecture, remains an unexplored area. In this work we aim to determine the impact of using chemical synapses compared to electrical synapses, in both sparse and all-to-all connected networks. We conduct experiments with autonomous lane-keeping through a photorealistic autonomous driving simulator to evaluate their performance under diverse conditions and in the presence of noise. The experiments highlight the substantial influence of the architectural and synaptic-model choices, respectively. Our results show that employing chemical synapses yields noticeable improvements compared to electrical synapses, and that NCPs lead to better results in both synaptic models. 

\end{abstract}

%%%%%%%%%%%%%%%%%%%%%%%%%%%%%%%%%%%%%%%%%%%%%%%%%%%%%%%%%%%%%%%%%%%%%%%%%%%%%%%%
\section{INTRODUCTION}

%Paragraph: About continuous-time recurrent neural networks
%\mathias{The very first sentence need to say that CT-RNNs are important for robotics and control (because it's ICRA, I already changed it)}
Continuous-time recurrent neural networks (CT-RNNs) are well-suited for robotic and control applications,
because they are able to learn dependencies and patterns occurring in CT processes~\cite{beer1992evolving,Funahashi1993ApproximationOD,neuralODEs,Hasani2020LiquidTN}. These networks are described by ordinary differential equations (ODEs) representing the time evolution of the activation of neurons over time, based on the inputs and the internal dynamics of the network. 

%Paragraph: why are they important, bio-inspired models
%TODO: add citations
Most robotic and control applications are safety-critical. Their faulty operation may cause human harm or physical damage. Interpretable machine learning methods are thus a natural fit for robotic applications, as they allow the system behavior to be studied and audited by humans. 

Liquid-time-constant neural networks (LTCs), a class of biology-inspired neural models with chemical synapses~\cite{Hasani2020LiquidTN}, have recently demonstrated interpretable-control capabilities in autonomous-driving and drone-navigation tasks~\cite{Lechner2020NeuralCP,Vorbach2021CausalNB, Kao2023}, when organized in a sparse 4-layer Neural Circuit Policy architecture (NCP)~\cite{worm_inspired}. However, the individual contributions of the NCP architecture and the synaptic model to interpretable control, is still poorly understood.

In this paper, we first show that electrical biologic synapses underlay CT-RNNs~\cite{Funahashi1993ApproximationOD}, and that chemical synapses underlay CT-RNNs with synapse-bound activations, and input-dependent forget gates~\cite{grnn}. We then study how much learning with electrical versus chemical synapses really makes a difference in the network's performance when they share the same type of architecture. For a systematic comparison, we experiment with LTCs~\cite{Hasani2020LiquidTN} and classic CT-RNNs~\cite{Funahashi1993ApproximationOD}, in both NCP~\cite{Lechner2020NeuralCP} and fully-connected architectures. 

We evaluate the performance of these networks in the complex task of autonomous car-navigation across different seasons. We first train the networks using imitation learning~\cite{Schaal1999IsIL}, where they learn to imitate the actions of a human driver through expert demonstrations. Subsequently, we deploy the trained networks in closed-loop settings to examine their adaptivity to previously unseen scenarios and their ability to react to environmental feedback. These evaluations are conducted using the data-driven photorealistic autonomous driving simulator VISTA~\cite{amini2022vista}. This has been proven to yield consistent results, when comparing trained agents in closed-loop evaluations in simulation, and real-world tests. 
%This makes it a highly reasonable choice due to its cost-effectiveness, safety and accessibility to the research community compared to conducting real car experiments. 

%Paragraph: what is our contribution here
Our main contributions are the following:
%\mathias{Don't summarize the experiment setup here but provide a more high level description of what you are studying, e.g., 1) We study the effect of the neuron model and wiring patterns of NCPs to better understand what ..... Point 3 should be "Our results show that ..." }
\begin{itemize}
    \item We show that saturated electrical synapses are the foundation of classic continuous-time RNNs, and that saturated chemical synapses underlay CT-RNNs with synapse-bound activations and input-dependent forget gates.  
    \item We conduct a comparative analysis between electrical and chemical synaptic models, respectively, in sparse NCP and fully-connected wiring architectures, evaluating their performance, interpretability and robustness.
    \item Our results show that: (1)~The architectural wiring choice significantly impacts the network performance, with NCPs exhibiting better results with both synaptic models, and (2)~Employing chemical synapses yields noticeable improvements compared to electrical synapses.
\end{itemize}

%\mathias{Maybe it would be good to add here very briefly how the paper is structured, e.g.. In section 2 we will recapitulate the neuron and wiring models of NCPs. In section 3, we provide a summary of related works on XXX. ...}
This paper is structured as follows: In Section~\ref{sec:synapticmodels} we discuss the bio-inspired models of electrical and chemical synapses, and the wiring model of NCPs. Section~\ref{sec:relatedwork} provides a summary of related work on imitation learning, lane-keeping, and applications of the models under investigation. In Section~\ref{sec:methods}, we outline the metrics of interest and detail the experimental setup. The results are presented in Section~\ref{sec:results}, followed by a concluding discussion in Section~\ref{sec:conclusion}.

\section{SYNAPTIC MODELS}\label{sec:synapticmodels}
The dynamic behavior of neurons is captured in neuroscience with electrical equivalent circuits (EECs)~\cite{wicks1996dynamic},~\cite{kandel2000principles}. In EECs, a neuron-$i$'s membrane is regarded as a capacitor whose potential $x_i$ is determined by the difference between intra- and extra-cellular concentrations of ions, respectively. The membrane potential thus evolves as follows:
\begin{equation}
    C\,\dot{x}_i = i_{li} + i_{si}, \quad i_{li} = g_{li}(e_{li} - x_i)
\label{eq:ltc}
\end{equation}
where $C$ is the membrane's capacitance, and $i_{li}$ and $i_{si}$ are the passive-leaking and dynamic-synaptic currents, traversing the capacitor, respectively. Without loss of generality, we will assume that $C\,{=}\,1$ in the rest of the paper. The leaking current, is determined by the Ohm's equation, where $e_{li}$ is the resting (or reversal) potential of the neuron, and $g_{li}$ is its leaking conductance. They are considered to be constant.

\vspace*{1mm}\subsubsection{Electrical synapses}
%Paragraph: Neuron models
%model the dynamic behavior of neurons
%These equations describe how the neural activity $x_i$ (membrane potential) changes over time in response to inputs $x_j$ (which can be external, from other neurons and self-connection).
%Intro to LTCs
occur in all organisms, especially in embryonic stage. Their prevalent form is that of an Ohmic current $i_{s,ji}\,{=}g_{ji}(y_j\,{-}\,x_i)$, where $y_j$ is either the potential $x_j$ of a presynaptic neuron, or the potential $u_j$ of an input. Constant $g_{ji}$ is the conductance of the synapse. Hence, in electrical synapses, $x_i$ faithfully follows $y_j$. Let $y\,{=}[x,u]$, $|x|\,{=}\,m$ and $|u|\,{=}\,n$. As a network, we thus obtain:
\begin{equation}
    \dot{x}_i =g_{li}(e_{li}-x_i) + \sum_{j=1}^{m+n} g_{ji} (y_j-x_i)
\label{eq:ES1}
\end{equation}
Rearranging this equation in terms of $x_i$ and $e_{li}$, one obtains:
\begin{equation}
    \dot{x}_i =-(g_{li}+\sum_{j=1}^{m+n} g_{ji})\,x_i + (g_{li}+\sum_{j=1}^{m+n} \frac{g_{ji}}{e_{li}} y_j)e_{li}
\label{eq:ES2}
\end{equation}
Equations~\eqref{eq:ES1}-\eqref{eq:ES2} are the EECs of neural networks with electrical synapses, only. However, neurons saturate their overall conductance, through diverse cellular mechanisms. A more faithful and numerically stable representation is thus:
\begin{equation}
    \dot{x}_i =-S(g_{li}+\sum_{j=1}^m g_{ji})\,x_i + T(g_{li}+\sum_{j=1}^m h_{ji}y_j)e_{li}
\label{eq:ctrnn}
\end{equation}
where $S$ and $T$ are sigmoidal and hyperbolic-tangent functions, respectively, and $h_{ji}\,{=}\,g_{ji}\,{/}\,e_{li}$. This equation is normalized, with conductances $g_{li},g_{ji}\,{\in}\,[0,1]$. Hence,  saturation is with $S$. Since $h_{ij}\,{\in}\,[-1,1]$ for $e_{li}\,{\in}\,[-1,1]$, we use $T$. 

Taking $\tau\,{=}\,S(g_{li}+\sum_{j=1}^m g_{ij})$ one obtains the classic CT-RNNs~\cite{Funahashi1993ApproximationOD}, with constant gate $\tau$. Thus, classic CT-RNNs are neural networks with electrical synapses, only!

\vspace*{1mm}\subsubsection{Chemical synapses}
occur in all organisms, especially in later stages. A current $i_{s,ji}\,{=}g_{ji}S(a_{ji}y_j+b_{ji})(e_{ji}\,{-}\,x_i)$ passes through a postsynaptic neuron's channels, when these open, as neurotransmitters released by the presynaptic neuron start binding to the receptors of the channels. Here, $g_{ji}$ is the maximum conductance of the channels, $S(a_{ji}y_j+b_{ji})$ is the probability of the channels to be open, with parameters $a_{ji}$ and $b_{ji}$, and $e_{ji}$ is the reversal potential of the channels. 

Note that in a chemical synapse, $x_i$ does not necessarily follow $y_j$. The synaptic current $i_{s,ji}$ is 0 even if all channels are open, when $e_{ji}\,{=}\,x_i$. A network is an ODEs system:
\begin{equation}
    \dot{x}_i =g_{li}(e_{li}-x_i) + \sum_{j=1}^{m+n} g_{ji}S(a_{ji}y_j+b_{ji})(e_{ji}-x_i)
\label{eq:CS1}
\end{equation}
These are the EECs (or LTCs~\cite{Hasani2020LiquidTN}) of chemical synapses. They also ignore saturation aspects. Taking those into account, and rearranging terms as before, one obtains SEECs (or SLTCs):
\begin{equation}
\label{eq:sltc}
\begin{array}{c}
    \dot{x}_i =-S(f_i)\,x_i + T(g_i)\,e_{li}\\[2mm]
    f_i = g_{li}+\sum_{j=1}^{m+n} g_{ji}S(a_{ji}y_j+b_{ji})\\[2mm]
    g_i = g_{li}+\sum_{j=1}^{m+n} h_{ji}S(a_{ji}y_j+b_{ji})
\end{array}
\end{equation}
As in gated RNNs~\cite{grnn}, $S(f_i)$ is the forget gate. However, the other gate is just $e_{li}$, the leaking potential. Moreover, each synapse has a different activation function (different $a_{ji}$, $b_{ji}$). 

\vspace*{1mm}\subsubsection{Neural Circuit Policies}
%Paragraph: NCP architecture and fully connected arch. 
for short NCPs, consist of 4 layers~\cite{worm_inspired}. The incoming information is received by the sensory neurons. This information is forwarded to interneurons, playing the role of a self-attention layer. Command neurons, which are connected in a recurrent fashion, store the state of the NCP, and use this together with the attention information, to compute actions. These are forwarded to the motor neurons. The NCP design aims to create a sparse architecture with interpretable dynamics. In contrast, commonly used fully-connected RNN architectures, do not distinguish layers of neurons, and neurons are connected all-to-all.

%\vspace*{1mm}\subsubsection{Numerical integration}
%CT-RNNs and SLTCs are often stiff ODEs, that require a very small integration step $dt$. By applying saturation, only after Euler expansion, considerably increases their stability. The difference equation is then:
%
%\begin{equation}
%\label{eq:sltc}
%    x_i(t+dt)=(1-S(f_i(t)\,dt)\,x_i(t) + T(g_i(t)\,dt)\,e_{li}
%\end{equation}
%
%This equation allows the ODE solver to employ considerably larger time increments $dt$, as the conductances $S(f_i(t)\,dt)$ and $T(g_i(t)\,dt)$ are still properly saturated.

\section{RELATED WORK}\label{sec:relatedwork}

%Paragraph: Imitation Learning
Imitation learning is about training the agents to learn the connections between perceptions and actions, by observing expert demonstrations of a task~\cite{Schaal1999IsIL}. Its application extends across various domains, including robotic manipulation~\cite{johns2021coarse}, locomotion~\cite{peng2020learning}, and autonomous navigation, both on the ground~\cite{Lechner2020NeuralCP, karnan2022voila} and in aerial environments~\cite{fan2020learn, Vorbach2021CausalNB, Kao2023}.
%

%Paragraph: End-to-end lane-keeping
Autonomous lane-keeping solutions include end-to-end convolutional neural networks (ConvNet)~\cite{Bojarski2016EndTE, Xu2016EndtoEndLO}, and more advanced solutions of ConvNet heads stacked with RNNs to consider previous information during decision making~\cite{Lechner2020NeuralCP}. Here, we use the latter variant, as similarly, we need a ConvNet head to preprocess the raw visual input which will be fed into recurrent chemical or electrical models.

%CTRNN and LTC 
The CT-RNN model of electrical synapses is a commonly used neural model, that has been applied in walking of a six-legged agent~\cite{beer1995dynamical}, robot-to-human handover~\cite{tang2016robot}, balancing a biped robot~\cite{Hnaff2020RealTI} and swarm robotics~\cite{sendra2023emergence}. Recently, the LTC model of chemical synapses has demonstrated the ability to autonomously navigate both cars~\cite{Lechner2020NeuralCP} and drones~\cite{Vorbach2021CausalNB, Kao2023}.
%TODO: add more examples to these

%Paragraph: Nature paper
%Previous work showed that using recurrent neural networks stacked with a convolutional head performs better than purely using end-to-end convolutional neural networks (CNN) in autonomous driving~\cite{Lechner2020NeuralCP}. 
An NCP wiring architecture built up from 19 LTC neurons, demonstrated better robustness compared to an end-to-end ConvNet, a fully-connected 64-neuron long-short term memory (LSTM)~\cite{hochreiter1997long} and a fully-connected 64-neuron CT-RNN~\cite{Lechner2020NeuralCP}. As this paper investigates autonomous lane-keeping too, our setup includes the two bio-inspired models from the mentioned paper: LTC neurons in a 19-neuron NCP architecture and 64 CT-RNNs all-to-all connected.  

%Paragraph: What is new here?
To the best of our knowledge, this paper is the first one to systematically explore the role of chemical and electrical synapses, in the same types of architectures. This combination enables us to provide a response to how much the architectural or synaptic model, respectively,  has an impact on the network's accuracy, robustness, and interpretability.
% Mixing network architectures and neural models
% Comprehensive evaluations of LNN using the VISTA simulator.

\section{METHODS}\label{sec:methods}
%Paragraph: Network setup: CNN+RNN

%TODO: write caption
%TODO: include layers in the image / in the caption
 \begin{figure*}[t]
    \centering    %\includegraphics[width=0.8\linewidth]{img/cnn_rnn3.pdf}
    \includegraphics[width=0.8\linewidth]{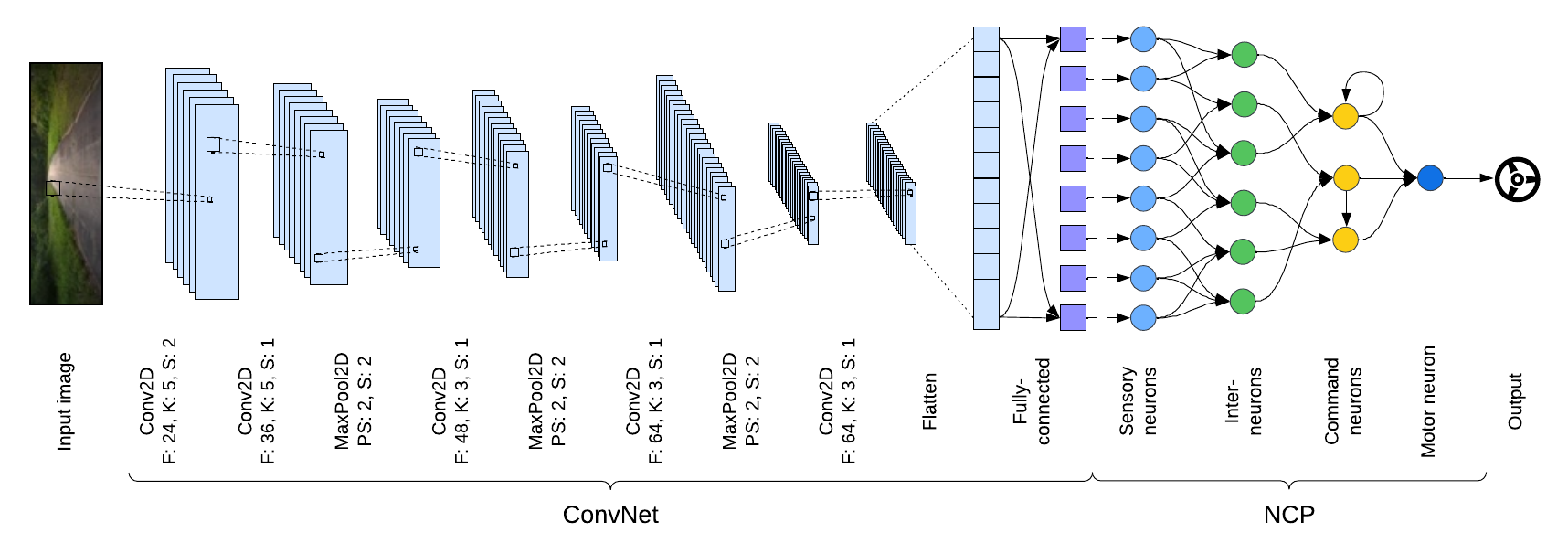} 
    \caption{The 4-layer NCP stacked with a convolutional head. The ConvNet extracts information from the input image which then serves as sensory input for the NCP. In a fully connected architecture, we replace the NCP with all-to-all wirings.}
     \label{fig:cnn_rnn}
     %\vspace*{-1ex}
\end{figure*}

%Paragraph: Imitation Learning
%Paragraph: VISTA simulator

%Experimental setup
We thoroughly analyze the impact of the architecture and the synaptic models by conducting two types of experiments:

\vspace*{1mm}\subsubsection{Fix the network architecture} In other words, change the synaptic model while keeping the number of neurons and their connections. This may result in an unfair performance comparison, because for the same network architecture, the simpler synaptic model used by CT-RNNs, offers much less trainable parameters. At the same time, the activity of the corresponding neurons is straightforward to compare.

\vspace*{1mm}\subsubsection{Fix the number of network parameters} In other words, change the network architecture, such that it results in the same number of trainable parameters. This way, the networks have an equal learning capacity. However, as CT-RNNs require a much bigger network architecture, they may lose the nice properties of interpretability and compactness. 
There is a plain trade-off between using the same network architecture, or the same number of trainable parameters.

As network bases, we use the LTC-NCP with 19 neurons, and the fully-connected version of CT-RNN with 64 neurons, as in~\cite{Lechner2020NeuralCP}. To close the knowledge gap between these two, we analyze the CT-RNNs in NCP architecture, with two sizes: 19 neurons, which corresponds to the type (1) experiment, and 64 neurons, which corresponds to the type (2) experiment, with approximately 1800 trainable parameters. Additionally, we include the LTCs in a fully-connected architecture, for a thorough analysis of the effect of the architecture. These setups are summarized in Table~\ref{tab:params}.

\subsection{Evaluation Metrics}
%TODO: check at the end if we listed everything here
We use several evaluation metrics to assess driving performance. They include prediction error, driving smoothness, and crash likelihood in closed-loop simulation on unknown trajectories. For interpretability, the metrics of interest are the network attention, and the changes in neuron activity during driving. In order to evaluate network robustness, in noisy conditions not present during training, we compare the attention maps and the driving ability, using the VISTA simulator. Our evaluation seeks to assess the importance of the synaptic model and the network architecture.

\begin{table}[tb]
\caption{Parameters of examined networks. The 19-neuron NCP has 12 interneurons, 6 command neurons and 1 motor neuron. The 64-neuron NCP is proportionally larger, with 42 interneurons, 21 command neurons and 1 motor neuron.}
\label{tab:params}
\begin{center}
\begin{tabular}{lccccc}
\hline
ID & \thead{Neural \\ model}  & \thead{Network \\ arch.} & \#Neuron & \#Synapse & \#Param\\ \hline
1 & LTC & NCP & 19 & 444 & 1833 \\ %\hline
2 & LTC & fully & 19 & 1577 & 6365 \\ %\hline
3 & CT-RNN & NCP & 19 & 444 & 482 \\ %\hline
4 & CT-RNN & NCP & 64 & 1700 & 1828  \\ %\hline
5 & CT-RNN & fully & 64 & 8192 & 8320 \\ \hline
\end{tabular}
\end{center}
\vspace*{-4ex}
\end{table}

\subsubsection{Training and Validation Loss}
We calculate the mean squared error (MSE) between the predicted road curvature $\hat{y_i}$ and the ground-truth $y_i$ of the validation set. Previous work~\cite{codevilla2018offline} pointed out that MSE alone, may not show the full driving quality. We thus use a second metric, weighting MSE values with the road steepness, at each time step. 

%\begin{equation}
%    MSE = a \frac{1}{|V|} \sum_{i \in V} (y_i-\hat{y_i})^2,
%\label{eq:mse}
%\end{equation}

%where $a$ is a scaling constant and $|V|$ is the number of samples in the validation set.

%Previous work~\cite{codevilla2018offline} pointed out that using the MSE might not show the full driving quality. Following this, we formulate a second error metric by weighting the MSE values: 

%\begin{equation}
%    MSE_W = a \frac{1}{|V|\cdot W} \sum_{i \in V} e^{(\alpha |\hat{y_i}|)} (y_i-\hat{y_i})^2
%\label{eq:weighted_mse}
%\end{equation}

%where the exponential term assigns higher weights to steeper road curvatures, $\alpha$ is a scaling constant and $W=\sum_{i \in V} e^{(\alpha |\hat{y_i}|)}$ normalizes the weights.

\vspace*{1mm}\subsubsection{Crash-Likelihood}
We aim to assess the performance of the networks in a closed-loop setting, where the next camera input is influenced by the previous steering decision. 
To this end, we measure the crash likelihood during multiple test runs. In the simulator, a crash is counted, if the car leaves the road.
To test the models in even more extreme conditions, we add extra Gaussian noise to the camera input, which was not present as a data augmentation technique in the training process. First, we use mean $\mu\,{=}\,0$ and variance $\sigma^2\,{=}\,0.1$, to see how they tackle this condition, and then we run experiments with doubled noise variance $\sigma^2\,{=}\,0.2$.

\vspace*{1mm}\subsubsection{Trajectory Smoothness}
%Lipschitz-constant
%Why + How to compute
To measure the smoothness of the driving trajectory, we compute the Lipschitz constant of the prediction over a test episode. A function $f$ is called Lipschitz continuous with Lipschitz constant $L\,{\geq}\,0$ if it satisfies: $|f(x_1)\,{-}\,f(x_2)|\,{\leq}\, L\,|x_1\,{-}\,x_2|$. If $|x_1\,{-}\,x_2|$ is the time passed between two timesteps, then $|f(x_1)\,{-}\,f(x_2)|$ is the corresponding change in the prediction. The value of $L$ measures the maximum rate of change in the network's output. Thus smaller values indicate smoother driving. 

\vspace*{1mm}\subsubsection{Saliency Maps and Structural-Similarity Index}
\begin{figure}[tb]
      \centering
      \includegraphics[width=0.9\linewidth]{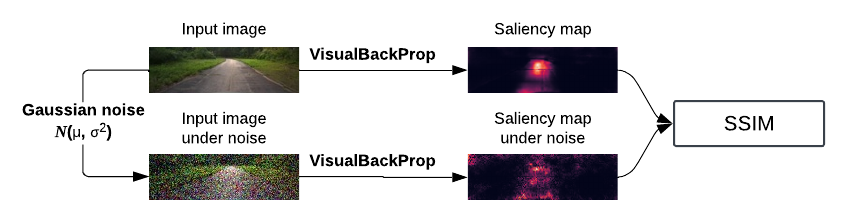}
      \caption{Structural-similarity index (SSIM) between saliency maps, under normal and Gaussian-noise conditions.}
      \label{fig:ssim}
   \end{figure}

%Visual backprop
%Why?
Getting insights into how a neural network makes its decision is very important in end-to-end autonomous systems. A possible approach is to analyze where the neural network pays the most attention during decision-making, that is,  find out which are the pixels in the input image that have the greatest impact. We observe the distinct evolution of the identical ConvNet architecture across all the models after training. 
%
%How?
To visualize the network's attention, we utilize the VisualBackprop algorithm~\cite{Bojarski2016VisualBackPropVC}, on our ConvNet heads. The main idea behind this method is, that the most relevant information comes from the feature maps of the deepest layers, which we bring backward layer by layer. This goes repeatedly, until the first layer, after which we take the absolute value of the attention, because the image standardization layer shifted the pixel values, and both positive and negative values should be considered with equal importance. 
%
%Algorithm~\ref{alg:visual_backprop} outlines the complete procedure. 
%
%\begin{algorithm}
%\caption{Generalized-VisualBackprop}
%\label{alg:visual_backprop}
%\begin{algorithmic}[1]
%\Require {$A = (A_0, A_1, ..., A_L)$, $A_i$: feature maps of the $i$-th layer of the ConvNet:  $L$: number of CNN layers}
%\Ensure {Saliency map}
%\Statex
%\Function{Generalized-VisualBackprop}{}
%\State $a \gets \text{avg}(A_L)$
%\For{$l \gets L-1$ to $0$}
%\State $f \gets \text{avg}(A_l)$
%\State $s \gets \text{size}(f)$
%\State $a \gets f \cdot \text{resize}(\text{map}=a,\text{new-size}=s)$
%\EndFor
%\State $a \gets \text{abs}(a)$
%\State \Return {$a$}
%\EndFunction
%\end{algorithmic}
%\end{algorithm}
%
%\subsubsection{Structural Similarity Index}
%Why?
To make one extra step, we aim to see the robustness of their learned ConvNet heads, in the presence of input noise, by comparing their saliency maps under normal and additional noise pairwise. The structural-similarity index (SSIM)~\cite{SSIM}, compares luminance, contrast, and structure of two signals, respectively. Signals are in our context images. In Figure~\ref{fig:ssim} we show how to compute the SSIM metric.

%The Structural Similarity Index (SSIM)~\cite{SSIM} between signals (in our context, images) represented by $\textbf{x}$ and $\textbf{y}$ is defined in the following way:

%\begin{equation}
%\begin{split}
%    \text{SSIM}(\textbf{x},\textbf{y}) & = [l(\textbf{x},\textbf{y})]^\alpha \cdot [c(\textbf{x},\textbf{y})]^\beta \cdot [s(\textbf{x},\textbf{y})]^\gamma \\
%    %& = \frac{(2 \mu_x \mu_y + C_1)(2 \sigma_{xy} + C_2)}{(\mu_x^2+ \mu_y^2+ C_1)(\sigma_x^2 + \sigma_y^2 + C_2)}
%\end{split}
%\label{eq:ssim}
%\end{equation}

%This index comprises three main components: $l$, $c$ and $s$, denoting luminance, contrast and structure comparisons, respectively. $\alpha>0$, $\beta>0$ and $\gamma>0$ alters the significance of these terms. In our experiments, we have assigned equal values of 1 to them. %For the assessment of luminance, $\mu_x$ and $\mu_y$ represent the mean intensities of the signals, while the standard deviations $\sigma_x$ and $\sigma_y$ serve as estimations for signal contrast. Additionally, $\sigma_{xy}$ denotes the correlation coefficient utilized in structure comparison. The constants $C_1$ and $C_2$ are defined as $C_1=(K_1L)^2$ and $C_2=(K_2L)^2$, where $K_1=0.01$ and $K_2=0.03$ are default parameter values of the algorithm, and $L=255$ is the dynamic range of pixel values in our specific scenario.

%SSIM: alpha, beta, gamma = 1, K1= 0.01, K2= 0.03, L=255

% About the results

\vspace*{1mm}\subsubsection{Neural Activity}
%Why and how
Another way to interpret, that is, to get insights into the network behavior, besides saliency maps, is to examine neural activity over time. Particularly focusing on whether the neural network's individual neurons exhibit correlated activity with straight road segments, and turns, along the driving path. This is achieved by extracting the driving trajectory from the simulator, as seen from a bird's-eye view, and then projecting the output of each neuron onto the car's trajectory, separately. To measure the similarity between the road curvature and the neural activity, we compute the cross-correlation of these signals. As we are interested in both negative and positive correlations, compared to no correlation, we take the absolute value of the correlations, and average them across all neurons in the network.

\subsection{Experimental Setup}

\begin{figure}[tb]
    \centering 
\begin{subfigure}{0.15\textwidth}
  \includegraphics[width=\linewidth]{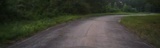}
  \label{fig:summer1}
\end{subfigure}\hfil 
\begin{subfigure}{0.15\textwidth}
  \includegraphics[width=\linewidth]{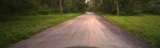}
  \label{fig:summer2}
\end{subfigure}\hfil
\begin{subfigure}{0.15\textwidth}
  \includegraphics[width=\linewidth]{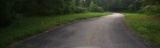}
  \label{fig:summer3}
\end{subfigure}
%\medskip
\begin{subfigure}{0.15\textwidth}
  \includegraphics[width=\linewidth]{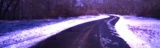}
 % \label{fig:winter1}
\end{subfigure}\hfil 
\begin{subfigure}{0.15\textwidth}
  \includegraphics[width=\linewidth]{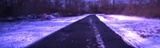}
\end{subfigure}\hfil
\begin{subfigure}{0.15\textwidth}
  \includegraphics[width=\linewidth]{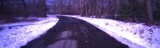}
\end{subfigure}
\caption{Examples from the dataset: summer and winter}
\label{fig:summer_winter_data}
\vspace*{-3ex}
\end{figure}

We evaluate the performance of the NCP and all-to-all architectures, with either electrical or chemical synapses, respectively, within an end-to-end autonomous driving scenario. Here the agent receives the front camera input of the car, and predicts the road curvature, which corresponds to the steering action of the lane-keeping task. Our dataset consists of human driving recordings, under different weather conditions~\cite{lechner2022all}. Figure~\ref{fig:summer_winter_data} shows samples from this dataset.

Predicting the road curvature instead of directly predicting the steering angle is beneficial because the curvature value is independent from the vehicle, while the steering angle is not. The steering angle can be computed from the curvature:
%
%From curvature to steering angle
\begin{equation}
    \alpha_t = S_v \: \textrm{arctan}(L_v \: y_t)
\end{equation}
where $\alpha_t$ is the steering angle at timestep $t$, $S_v$ and $L_v$ are the steering-ratio and wheelbase properties of the vehicle, respectively, and $y_t$ is the road curvature at timestep $t$.

 \begin{figure}[tb]
  \centering
  %\framebox{\parbox{3in}{We suggest that you use a text box to insert a graphic (which is ideally a 300 dpi TIFF or EPS file, with all fonts embedded) because, in an document, this method is somewhat more stable than directly inserting a picture.
%}}
  \includegraphics[width=0.9\linewidth]{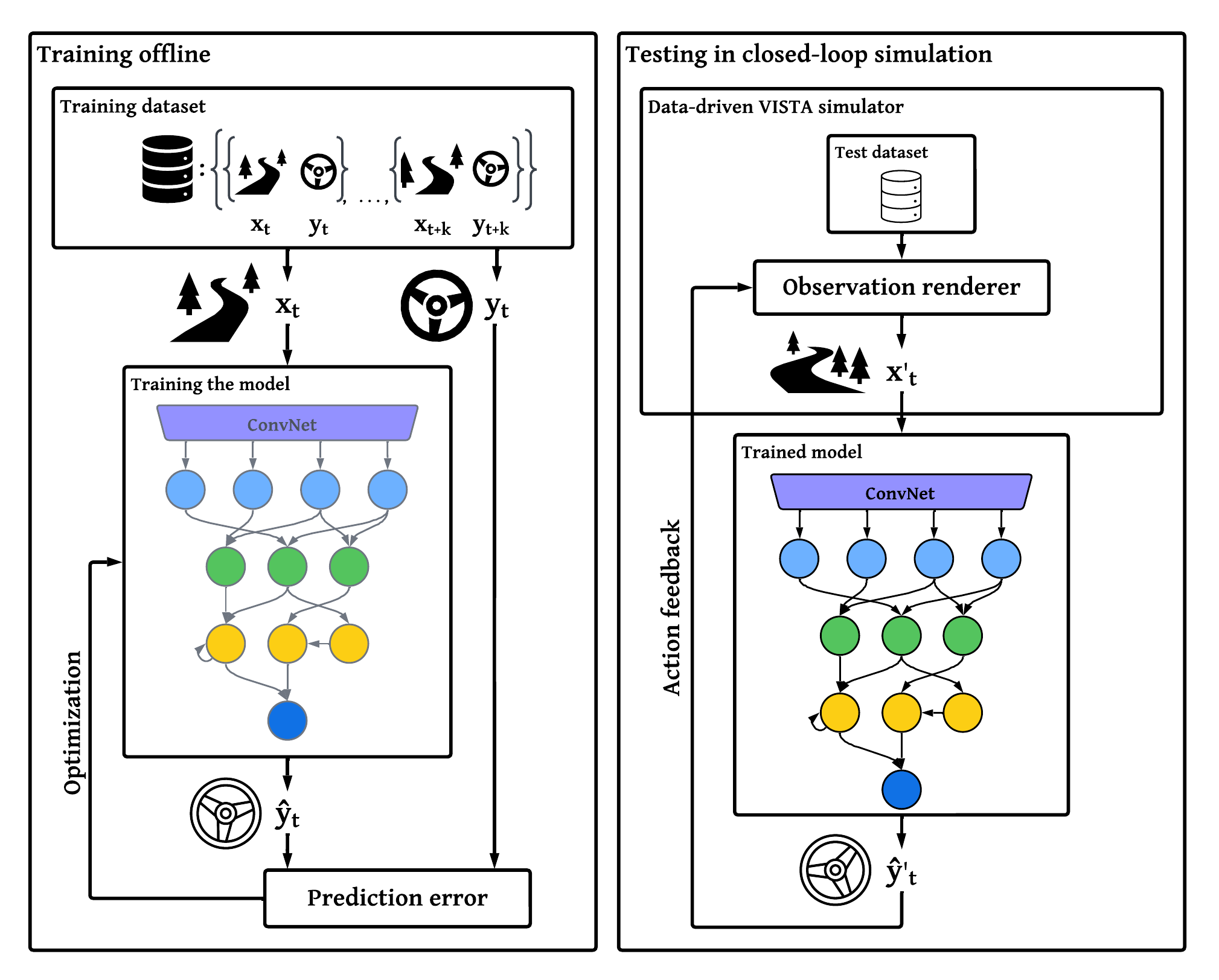}
  \caption{Training and testing procedure.}
  \label{fig:training_test}
  \vspace*{-2ex}
\end{figure}

%Open-loop
For training purposes, we create an open-loop imitation learning task, from summer and winter data.  The training data contains input-output sequences $\{(x_t, y_t)\,|\,t\,{\in}[0..T]\}$, where $x_t$ is the camera frame, and $y_t$ is the road curvature at time $t$. The network receives the camera frame $x_t$ at time step $t$ and predicts the road curvature $\hat{y}_t$. 

%Preprocessing
As data augmentation, we use the following procedure. We first adjust the brightness, by adding to the pixels a random value chosen from a uniform distribution in $[-0.4, 0.4]$. After this, the contrast and saturation of the image are shifted, by multiplying each RGB channel by a contrast factor, and the saturation channel in its HSV representation by a saturation factor, respectively. These random values are taken from a uniformly distribution in $[0.6,1.4]$. In addition to this, the training data is augmented by Guided Policy Learning~\cite{pmlr-v28-levine13} using the VISTA simulator~\cite{Amini2020LearningRC, amini2022vista}. This places the car off-orientation on the track, and the model has to learn to recover the car from these quite challenging positions.

%About training procedure
%table: hyperparameters
The model is trained using the mean squared error (MSE) of the prediction $\hat{y}_t$ and the label $y_t$. This is offline learning: the prediction of the network does not influence the next input image of the sequence. This can be seen on the left-hand side in Figure~\ref{fig:training_test}. We run the training on a sequence of input images of length 32, with a batch size of 32, with a cosine learning-rate decay schedule with an initial value of $10^{-3}$ over 100 epochs. We apply the AdamW optimizer~\cite{Loshchilov2017FixingWD} which in our case decouples the weight decay with the value of $10^{-6}$ from the learning rate. The best epoch is determined by the highest validation loss, without early stopping, in order to achieve a better generalization~\cite{Power2022GrokkingGB}.

%table: conv head layers

After training the models, we test them in the data-driven VISTA simulation platform, which synthesizes new possible trajectories from the dataset. This way we can evaluate the generalization performance of the models, in a closed-loop environment, where the prediction of the current time step affects the camera input of the next time step. This approach is illustrated on the right-hand side in Figure~\ref{fig:training_test}.

\section{RESULTS}\label{sec:results}

\subsection{MSE in Open-Loop}

Table~\ref{tab:val_loss} compares the neural networks in terms of MSE and Weighted MSE in the open-loop setting. We observe that the CT-RNN NCP-64 model exhibited the lowest MSE error, indicating the best fit to the data compared to the other models. By assigning higher weights to predictions in steeper curves, LTC-based models showed better performance to it by achieving the lowest Weighted MSE values.

\begin{table}[tb]
\caption{Prediction error in the validation set, for 5 trained models, per model-type. Smaller error is better performance.}
\label{tab:val_loss}
\begin{center}
\begin{tabular}{lccc}
\hline
Model               & MSE & Weighted MSE\\ \hline
LTC NCP-19        & 0.173 $\pm$ 0.021 & \textbf{0.011 $\pm$ 0.001} \\ %\hline
LTC fully-19      &  0.200 $\pm$ 0.027 & \textbf{0.010 $\pm$ 0.001} \\ %\hline
CT-RNN NCP-19  & 0.174 $\pm$ 0.012 & 0.015 $\pm$ 0.002 \\ %\hline
CT-RNN NCP-64   & \textbf{0.163 $\pm$ 0.017} & \textbf{0.012 $\pm$ 0.002} \\ %\hline
CT-RNN fully-64 & 0.219 $\pm$ 0.015 & 0.018 $\pm$ 0.002 \\ \hline
\end{tabular}
\end{center}
\vspace*{-3ex}
\end{table}

%TODO: include a nice plot of steering angle prediction over time to have more of an icra style paper - sketch done
%Next: choose 2 models, that have similar mse but different angle-weighted mse values --- replace images
% \begin{figure}[tb]
%  \centering
%  \includegraphics[width=0.9\linewidth]{img/offline_prediction_subplots.pdf}
%  \caption{Evaluation of XYZ models. --placeholder-- }
%  \label{figurelabel}
%\end{figure}

\subsection{Crash-Likelihood in Closed-Loop}

% season x (5x25) =  2  x 125 runs per model
We tested the networks for 125 episodes for each season and calculated how likely they drive off the road over these episodes. We found that the chemical-synapse-based LTCs independently of the architecture, and the electrical-synapse-based CT-RNNs in NCP wiring architectures drive without crash in the closed-loop setting using the VISTA simulator.  
Moreover, they are robust enough, to tackle the challenge of receiving input images with 0-mean and $0.1$-variance Gaussian noise. In contrast, the fully-connected CT-RNN network crashes without any input disturbances during winter and under additional noise in both summer and winter tests. 

We also aimed to push these models to their limits, and observed that by doubling the noise variance, the models are not able to avoid the crashes anymore. The trend of the crash likelihood shows that the 4-layer structure of the NCP brings an advantage compared to the fully-connected variant.

Note that some of the evaluation metrics of the CT-RNN in fully-connected architecture could be not relevant, as this was not able to drive without crashes under normal circumstances during winter, and under 0.1-variance Gaussian noise in both seasons. However, we included them in our full analysis.

\begin{table}[tb]

\caption{Crash likelihood in closed-loop simulation, under noise-free conditions, and under Gaussian noise with mean of 0 and variance of 0.1 and 0.2. Smaller is better. S is summer, W is winter. 125 runs per model in each condition.}
\label{tab:crash_all}
\begin{center}
\begin{tabular}{lcccccc}
\hline
 & \multicolumn{2}{c}{Noise-free} & \multicolumn{2}{c}{$\sigma^2=0.1$} & \multicolumn{2}{c}{$\sigma^2=0.2$} \\ 
Model & S  & W  & S  & W & S  & W \\ \hline
LTC NCP-19         & \textbf{0.0} & \textbf{0.0} & \textbf{0.0} & \textbf{0.0} & 0.11 & \textbf{0.01}\\ %\hline
LTC fully-19      & \textbf{0.0} & \textbf{0.0} & \textbf{0.0} & \textbf{0.0} & 0.18 & 0.02 \\ %\hline
CT-RNN NCP-19  & \textbf{0.0} & \textbf{0.0}  & \textbf{0.0} & \textbf{0.0} & 0.20  & 0.08\\ %\hline
CT-RNN NCP-64   & \textbf{0.0} & \textbf{0.0}  & \textbf{0.0} & \textbf{0.0}  & \textbf{0.05}  & 0.03 \\ %\hline
CT-RNN fully-64 & \textbf{0.0} & 0.02 & 0.09 & 0.06  & 0.68  & 0.25\\ \hline
\end{tabular}
\end{center}
\end{table}

\subsection{Trajectory Smoothness}

Table~\ref{tab:lipschitz_simtrajectory} provides the Lipschitz constants for networks in summer and winter contexts, both in noise-free conditions and under the influence of 0-mean and 0.1-variance noise. Observe that LTC networks generally exhibit smoother driving performance compared to CT-RNNs, in most scenarios. Moreover, when employing the NCP architecture, LTCs achieve an exceptionally smooth trajectory.

%TODO: create a figure instead of this, to fit into 1 column
\begin{table*}[tb]
\caption{Lipschitz constant of the prediction in the first 4 columns. Smaller is better. Similarity of the trajectory between normal conditions and under Gaussian noise variance of 0.1 in the last 2 columns. Higher is better. LTC-based networks can maintain a smooth driving trajectory in the presence of noise better than CT-RNN networks. Using an NCP architecture makes the neural network more resilient. Note that even though the CT-RNN fully-64 results show good characteristics, this model drives off the road under this amount of noise, while the others can drive without crashes.}
\label{tab:lipschitz_simtrajectory}
\begin{center}
\begin{tabular}{lcccc|cc}
\hline
 & \multicolumn{4}{c|}{Lipschitz constant of the output}  & \multicolumn{2}{c}{Similarity of the trajectory}                                                        \\
 & \multicolumn{2}{c}{Noise-free} & \multicolumn{2}{c|}{Under noise $N(\mu=0, \sigma^2=0.1)$}  & \multicolumn{2}{c}{Under noise $N(\mu=0, \sigma^2=0.1)$}                                                        \\
Model  & Summer  & Winter  & Summer  & Winter & Summer  & Winter \\ \hline
LTC NCP-19        & 0.0081 $\pm$ 0.0028  & \textbf{0.0087 $\pm$ 0.0031} & \textbf{0.0115 $\pm$ 0.0028} & \textbf{0.0113 $\pm$ 0.0036} & \textbf{0.971 $\pm$ 0.043} & \textbf{0.980 $\pm$0.027}\\ %\hline
LTC fully-19      & \textbf{0.0076 $\pm$ 0.0022}  & 0.0101 $\pm$ 0.0045 & 0.0124 $\pm$ 0.0028  & 0.0126 $\pm$ 0.0041 & 0.967 $\pm$ 0.046 & 0.976 $\pm$ 0.030 \\ %\hline
CT-RNN NCP-19  & 0.0107 $\pm$ 0.0097 & 0.0150 $\pm$ 0.0098 & 0.0143 $\pm$ 0.0093 & 0.0172 $\pm$ 0.0092  & 0.964 $\pm$ 0.054 &  0.968 $\pm$ 0.047 \\ %\hline
CT-RNN NCP-64   & 0.0109 $\pm$ 0.0067 & 0.0124 $\pm$ 0.0070 & 0.0124 $\pm$ 0.0060 & 0.0148 $\pm$ 0.0070 & 0.962 $\pm$ 0.057 &  0.967 $\pm$ 0.047\\ %\hline
CT-RNN fully-64 & \textbf{0.0076 $\pm$ 0.0029}  & 0.0107 $\pm$ 0.0089 & 0.0126 $\pm$ 0.0033  & 0.0122 $\pm$ 0.0049 & 0.911 $\pm$ 0.171 & 0.970 $\pm$ 0.040 \\ \hline
\end{tabular}
\end{center}
\end{table*}

\subsection{Network Interpretability}

%TODO: include 1 summer, 1 winter saliency maps
\subsubsection{Saliency maps} 
\begin{figure}[tb]
\vspace*{3ex}
    \centering 

\begin{subfigure}{0.15\linewidth}
  \includegraphics[width=\linewidth]{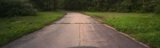}
    %\caption{}
  \label{fig:frame_summer}
\end{subfigure}
\begin{subfigure}{0.15\linewidth}
  \includegraphics[width=\linewidth]{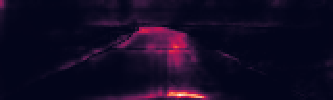}
  %\caption{}
  \label{fig:ltc_ncp19_saliency_summer}
\end{subfigure}\hfil 
\begin{subfigure}{0.15\linewidth}
  \includegraphics[width=\linewidth]{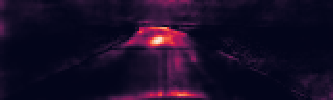}
   %\caption{}
  \label{fig:ltc_fully19_saliency_summer}
\end{subfigure}\hfil
\begin{subfigure}{0.15\linewidth}
  \includegraphics[width=\linewidth]{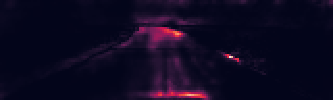}
   %\caption{}
  \label{fig:ctrnn_ncp19_saliency_summer}
\end{subfigure}\hfil
\begin{subfigure}{0.15\linewidth}
  \includegraphics[width=\linewidth]{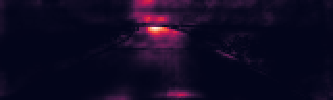}
   %\caption{}
  \label{fig:ctrnn_ncp64_saliency_summer}
\end{subfigure}\hfil
\begin{subfigure}{0.15\linewidth}
  \includegraphics[width=\linewidth]{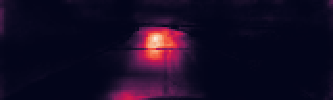}
   %\caption{}
  \label{fig:ctrnn_fully64_saliency_summer}
\end{subfigure}

\begin{subfigure}{0.15\linewidth}
  \includegraphics[width=\linewidth]{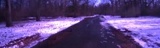}
    \caption{}
  \label{fig:frame_winter}
\end{subfigure}
\begin{subfigure}{0.15\linewidth}
  \includegraphics[width=\linewidth]{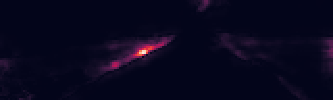}
  \caption{}
  \label{fig:ltc_ncp19_saliency}
\end{subfigure}\hfil 
\begin{subfigure}{0.15\linewidth}
  \includegraphics[width=\linewidth]{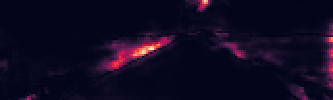}
   \caption{}
  \label{fig:ltc_fully19_saliency}
\end{subfigure}\hfil
\begin{subfigure}{0.15\linewidth}
  \includegraphics[width=\linewidth]{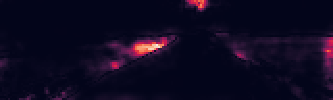}
   \caption{}
  \label{fig:ctrnn_ncp19_saliency}
\end{subfigure}\hfil
\begin{subfigure}{0.15\linewidth}
  \includegraphics[width=\linewidth]{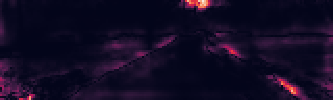}
   \caption{}
  \label{fig:ctrnn_ncp64_saliency}
\end{subfigure}\hfil
\begin{subfigure}{0.15\linewidth}
  \includegraphics[width=\linewidth]{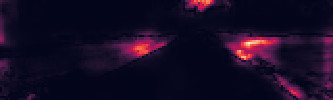}
   \caption{}
  \label{fig:ctrnn_fully64_saliency}
\end{subfigure}

\caption{Summer-winter input images in (a), and their saliency maps of LTC NCP-19 in (b), LTC fully-19 in (c), CT-RNN NCP-19 in (d), CT-RNN NCP-64 in (e) and CT-RNN fully-64 in (f). Observe that an NCP architecture makes the models more focused compared to their fully connected version. The LTC NCP-19 keeps its attention on the road and its contours in summer, and on the road contours in winter. It ignores the irrelevant information from the border of the image.}
\label{fig:saliency}
\vspace*{-1ex{}}
\end{figure}

Figure~\ref{fig:saliency} shows the saliency maps of the networks considered, for both summer and winter. We found that the networks extract information differently in the two seasons: the attention goes to the road itself in summer, and the contour (the side) of the road in winter. Intuitively, the areas of the input image containing the most relevant steering information should be highlighted: the road, and its contours, without the unnecessary environmental features which occur outside of the road. Note that the NCPs focus on the relevant parts, alone, while the fully connected networks also consider irrelevant areas of the input frames.  

%Neural activity during driving
\vspace*{1mm}\subsubsection{Neural Activity} 
\begin{figure}[tb]
    \centering 
\begin{subfigure}{0.15\linewidth}
  \includegraphics[width=\linewidth]{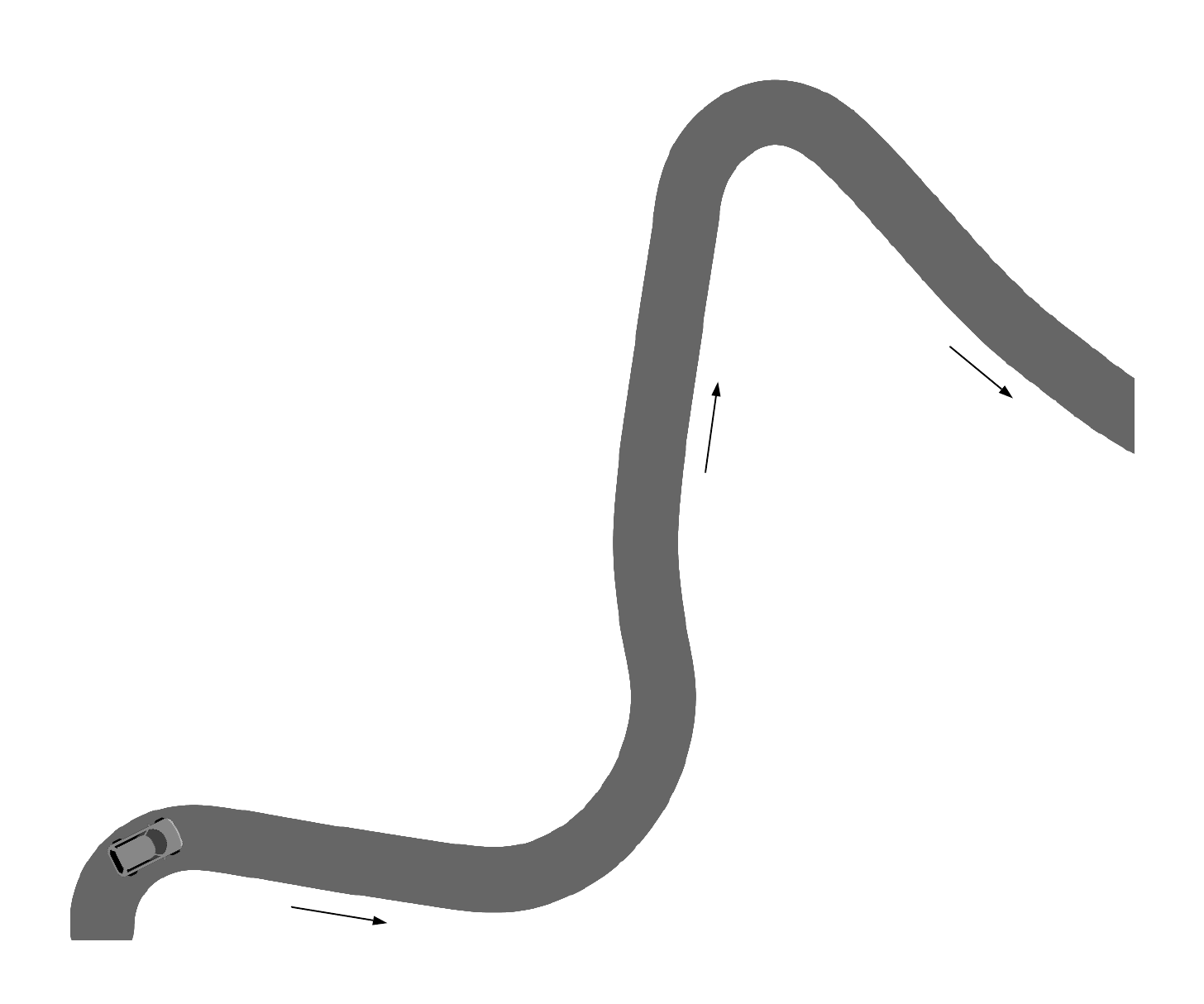}
    \caption{}
  \label{fig:road_curv}
\end{subfigure}
\begin{subfigure}{0.15\linewidth}
  \includegraphics[width=\linewidth]{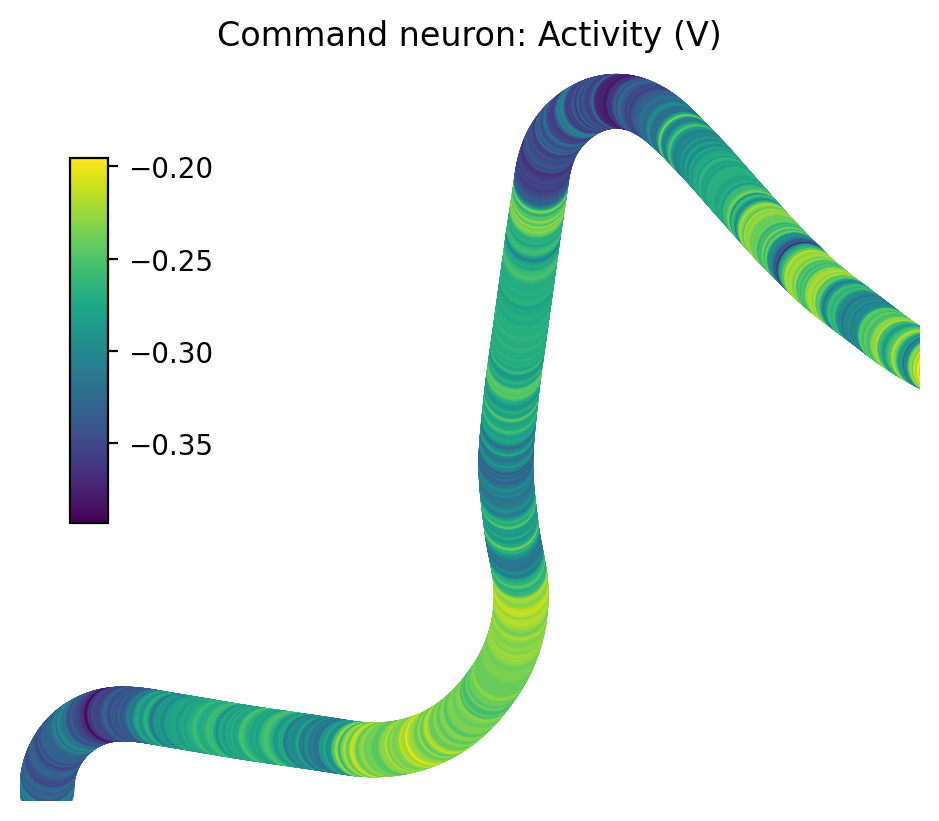}
  \caption{}
  \label{fig:ltc_ncp19_activity}
\end{subfigure}\hfil 
\begin{subfigure}{0.15\linewidth}
  \includegraphics[width=\linewidth]{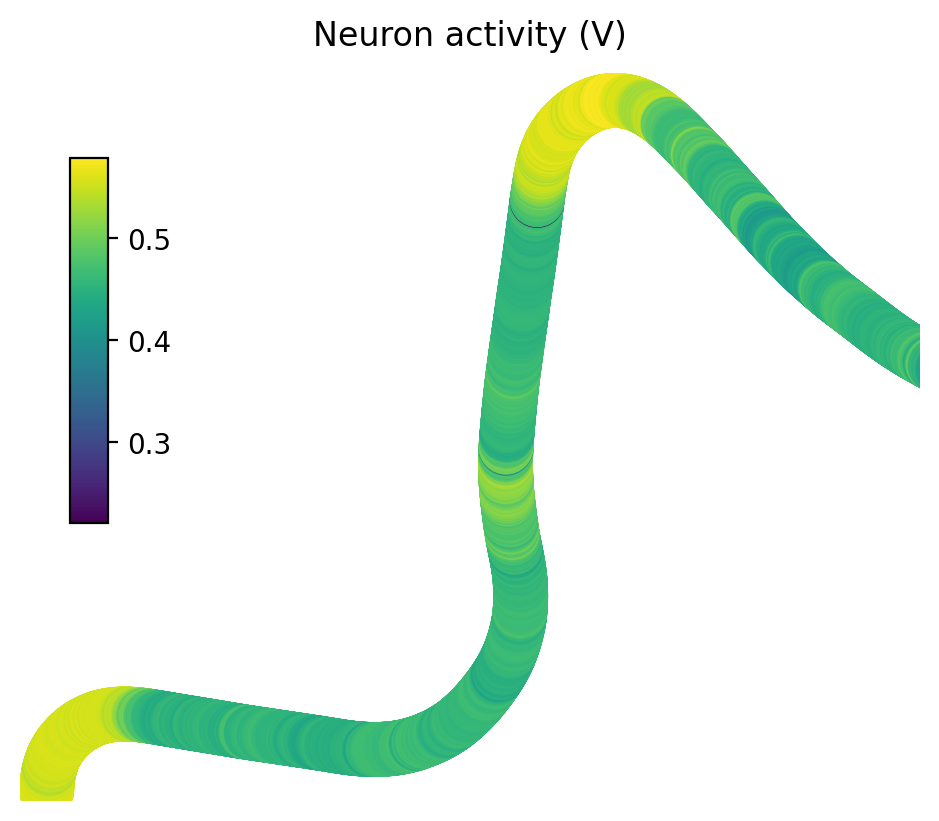}
   \caption{}
  \label{fig:ltc_fully19_activity}
\end{subfigure}\hfil
\begin{subfigure}{0.15\linewidth}
  \includegraphics[width=\linewidth]{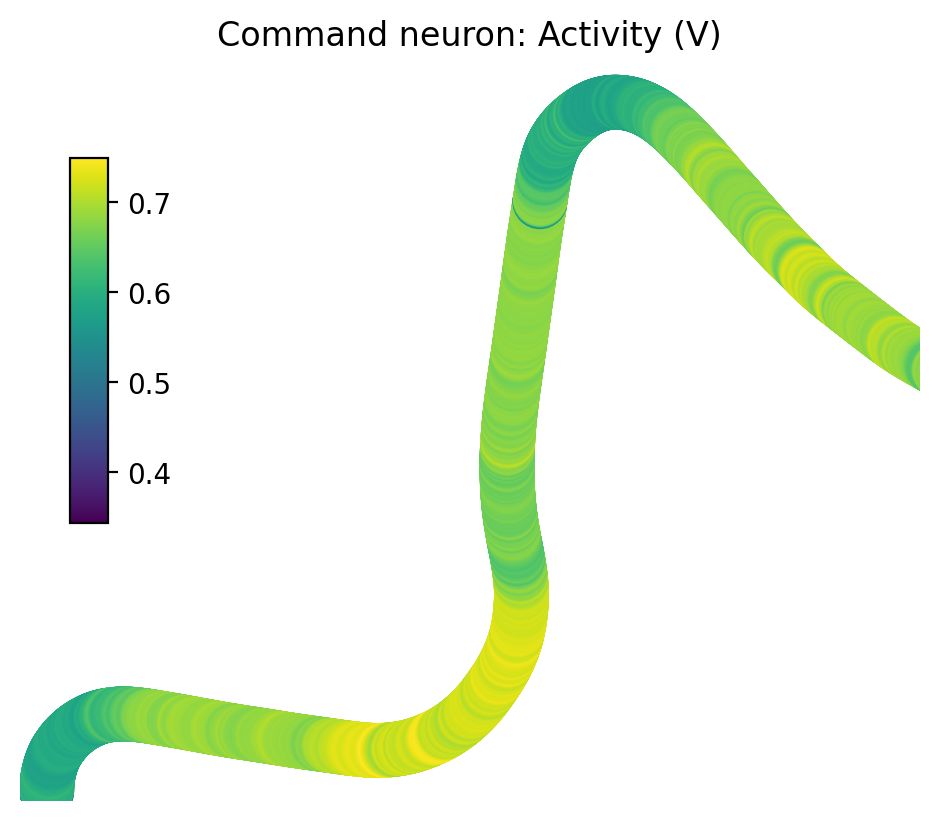}
   \caption{}
  \label{fig:ctrnn_ncp19_activity}
\end{subfigure}\hfil
\begin{subfigure}{0.15\linewidth}
  \includegraphics[width=\linewidth]{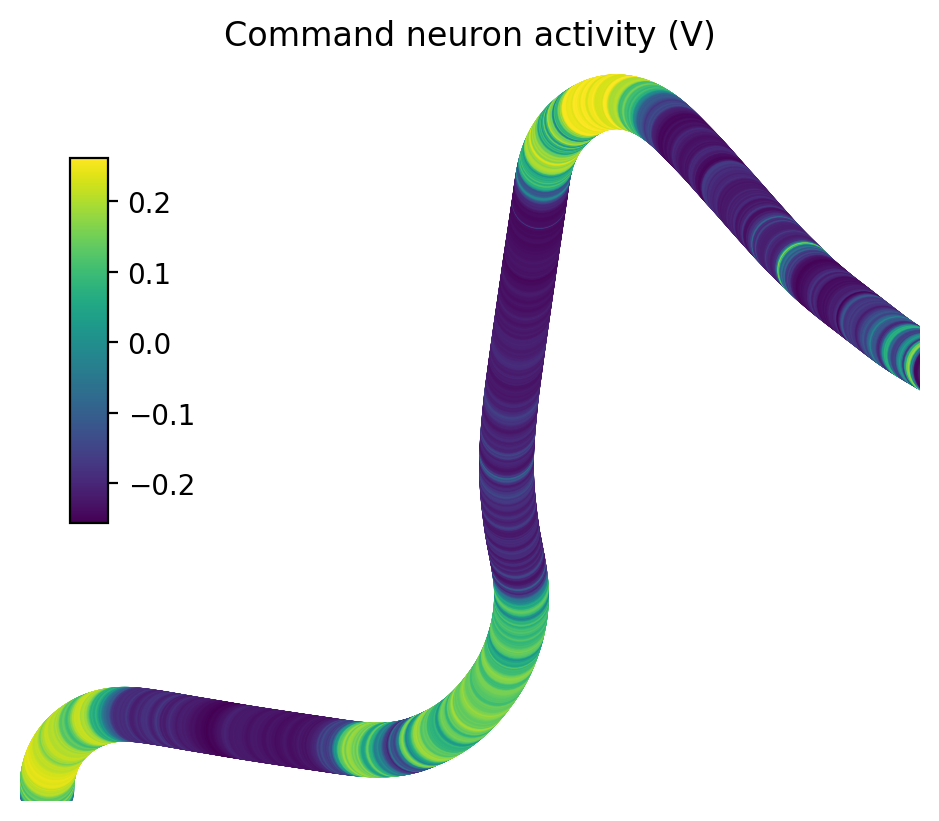}
   \caption{}
  \label{fig:ctrnn_ncp64_activity}
\end{subfigure}\hfil
\begin{subfigure}{0.15\linewidth}
  \includegraphics[width=\linewidth]{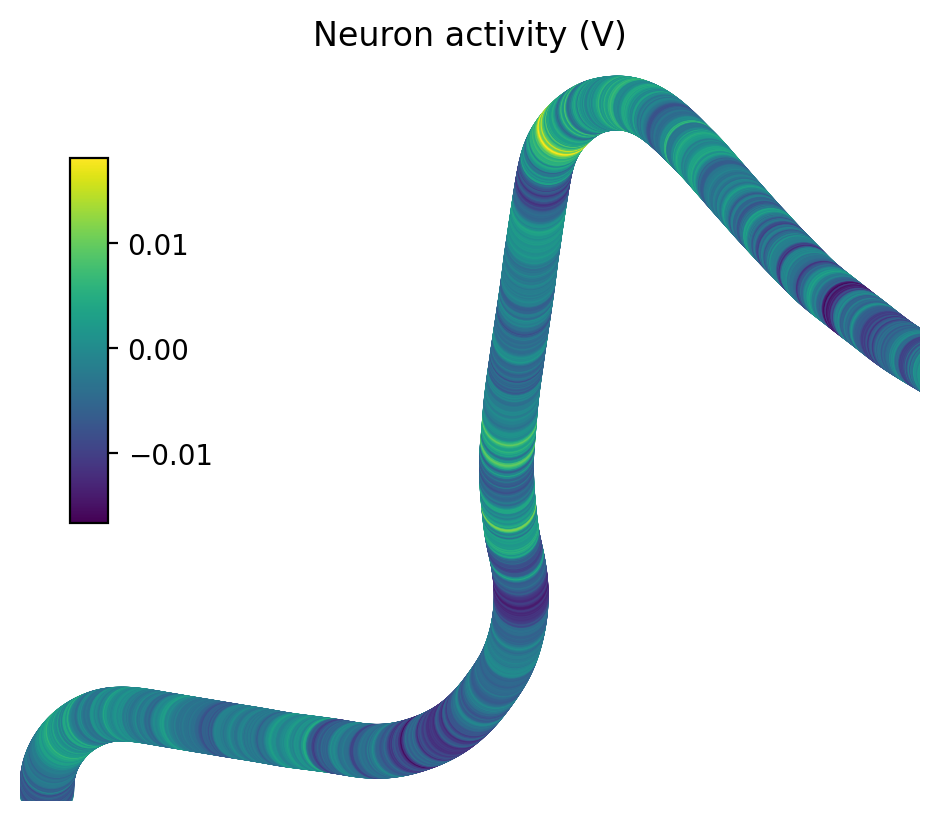}
   \caption{}
  \label{fig:ctrnn_fully64_activity}
\end{subfigure}

\caption{Neural activity. (a) Road trajectory from above, arrows indicate the driving direction (b)-(f)  Models in the same order as before. The NCP architecture brings interpretability to CT-RNNs as the changes in neural activity correspond to the road's trajectory, as illustrated in (d)-(e). LTC in all-to-all architecture is still interpretable in this sense, as seen in (c).}
\label{fig:neural_activity}
\end{figure}

When considering the neural activity during driving, we observed that in NCP architectures,
neural activities change their dynamics based on the road curvature, regardless of the synaptic type. This behavior holds for LTCs in the fully-connected architecture, too, which means that this interpretability feature holds regardless of the network architecture, for chemical synapses. However, the electrical-synapse-based CT-RNN does not have this nice property without the NCP structure. Its all-to-all wiring shows no patterns of different activation during turns and straight sections of the road. For illustration, Figure~\ref{fig:neural_activity} shows the activity of one of the neurons in the examined networks.
In addition, we also computed the average correlation between the neural activity of each neuron, and the driving trajectory. The results in Table~\ref{tab:corr_activity} clearly show that the LTC neurons have the best interpretability in terms of network dynamics, irrespective of the architecture used. In contrast, for the classic CT-RNN models, the NCP architecture gets around 1.5-2 times increased activity correlation compared to the all-to-all one. Hence, the NCP architecture brings CT-RNNs very close to the chemical-synapse-based LTC model.

\begin{table}[tb]
\caption{Correlation between the neural activity and the driving trajectory. Higher is better in this case. The NCP architecture brings the CT-RNN model much closer to the LTC-type networks, in terms of interpretable neural activity.}
\label{tab:corr_activity}
\begin{center}
\begin{tabular}{lcc}
\hline
Model               & Summer & Winter \\ \hline
LTC NCP-19        & \textbf{0.662 $\pm$ 0.290} & 0.636 $\pm$ 0.279 \\ %\hline
LTC fully-19      & \textbf{0.661 $\pm$ 0.307}  & \textbf{0.698 $\pm$ 0.270} \\ %\hline
CT-RNN NCP-19  & 0.649 $\pm$ 0.316 & 0.630 $\pm$ 0.293 \\ %\hline
CT-RNN NCP-64   & 0.568 $\pm$ 0.313 & 0.524 $\pm$ 0.282  \\ %\hline
CT-RNN fully-64 & 0.395 $\pm$ 0.282  & 0.321 $\pm$ 0.254 \\ \hline
\end{tabular}
\end{center}
\end{table}
\vspace*{-1ex}

\subsection{Network Robustness}

\begin{figure}[tb]
  \centering
  \includegraphics[width=0.49\linewidth]{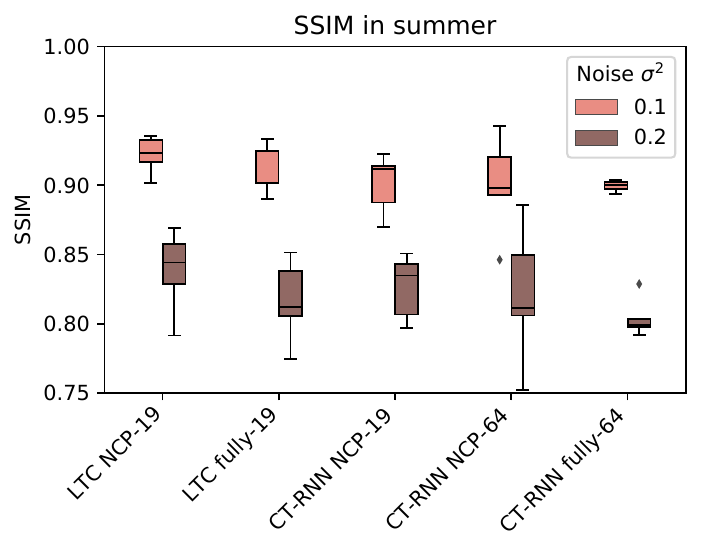}
  \includegraphics[width=0.49\linewidth]{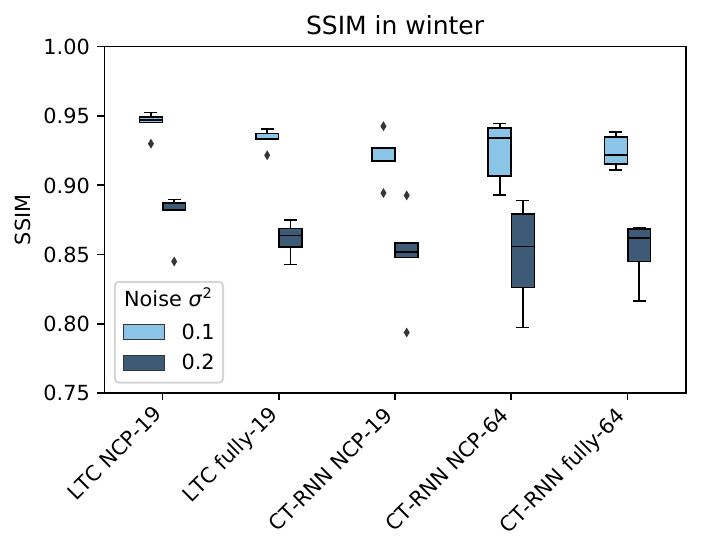} 
  \caption{The structural-similarity index (SSIM) for the neural networks considered, under Gaussian noise with mean 0, and variance 0.1 and 0.2, in summer on the left-hand side and winter on the right-hand side. Higher is better.}
  \label{fig:ssim_plots}
\end{figure}

Finally, we measured how much the saliency maps of the various networks, changed in the presence of a noisy input. An SSIM value close to 1 indicates a more robust attention mechanism, meaning that the additional noise has less of an influence on the convolutional head of these networks. In Figure~\ref{fig:ssim_plots} we present our results. These show that the LTC model in the NCP wiring architecture is the most robust: it extracts nearly identical information from the input images even when noise is present.

In order to evaluate how well the five models considered can maintain their driving trajectory, we also calculated the similarity between their trajectories with and without noise. These results are reported and discussed in Table~\ref{tab:lipschitz_simtrajectory}.

%\begin{table}[tb]
%\caption{Similarity of the curvature under normal conditions and under Gaussian noise variance of 0.1 (higher is better). LTC-based networks can maintain the same driving trajectory in the presence of noise better than CTRNN networks. Using NCP architecture makes the models more resilient.}
%\label{tab:corr_noise}
%\begin{center}
%\begin{tabular}{lcc}
%\hline
%Model & Summer & Winter    \\ \hline
%LTC NCP-19 & \textbf{0.971 $\pm$ 0.043} & \textbf{0.980 $\pm$0.027}         \\ %\hline
%LTC fully-19 & 0.967 $\pm$ 0.046 & 0.976 $\pm$ 0.030 \\ %\hline
%CTRNN NCP-19   & 0.964 $\pm$ 0.054 &  0.968 $\pm$ 0.047         \\ %\hline
%CTRNN NCP-64  & 0.962 $\pm$ 0.057 &  0.967 $\pm$ 0.047       \\ %\hline
%CTRNN fully-64 & 0.911 $\pm$ 0.171 & 0.970 $\pm$ 0.040 \\ \hline
%\end{tabular}
%\end{center}
%\end{table}

%\section{CONCLUSIONS AND FUTURE WORK}\label{sec:conclusion}
\section{CONCLUSIONS}\label{sec:conclusion}
In this paper, we showed how neuroscience models of electrical and chemical synapses result in continuous-time recurrent neural networks. These bio-inspired neural networks hold the potential for interpretable solutions, which are highly sought after in robotics systems. 

We investigated the impact of synaptic choices in the architectures of sparse and fully-connected wirings. Through extensive experiments conducted in a photorealistic autonomous driving simulator, we evaluated their performance, interpretability, and robustness under diverse conditions and in the presence of noise.

Our findings revealed that the choice of wiring architecture had a significant impact on the networks' performance. Specifically, the sparse NCP wirings were more effective with both synaptic models. Concerning the synaptic models themselves, our research demonstrated that learning with chemical synapses makes a difference. They not only led to a more robust behavior but also resulted in better interpretability and significantly smoother driving dynamics when compared to electrical synapses.

%future work if there's enough space left

%%%%%%%%%%%%%%%%%%%%%%%%%%%%%%%%%%%%%%%%%%%%%%%%%%%%%%%%%%%%%%%%%%%%%%%%%%%%%%%%
%\section*{APPENDIX}

%\section*{ACKNOWLEDGMENT}
%
%TODO: Include the text that Ioanna sent 
%This project has received funding from the European Union’s Horizon 2020 research and innovation programme under the Marie Skłodowska-Curie grant agreement No 101034277.

%%%%%%%%%%%%%%%%%%%%%%%%%%%%%%%%%%%%%%%%%%%%%%%%%%%%%%%%%%%%%%%%%%%%%%%%%%%%%%%%

\bibliography{refs}
\bibliographystyle{ieeetr}

\end{document}